\documentclass{article}
\usepackage{ijcai19}

\usepackage[OT1]{fontenc} 
\usepackage{times}
\usepackage{soul}
\usepackage{url}
\usepackage[hidelinks]{hyperref}
\usepackage[utf8]{inputenc}
\usepackage[small]{caption}
\usepackage{graphicx}
\usepackage{amsmath}
\usepackage{booktabs}
\usepackage{algorithm}
\usepackage{algorithmic}
\usepackage{enumerate}
\usepackage{color}
\usepackage{multirow}
\usepackage{hyperref}
\usepackage{diagbox}
\usepackage{threeparttable}
\newcommand{\eg}{\emph{e.g.}}
\newcommand{\ie}{\emph{i.e.}}
\newcommand{\etc}{\emph{etc}}

\title{Learning Instance-wise Sparsity for Accelerating Deep Models}

\author{
Chuanjian Liu$^1$\footnote{Contact Author}\and
Yunhe Wang$^1$\and
Kai Han$^1$\and
Chunjing Xu$^1$\And
Chang Xu$^2$\\
\affiliations
$^1$Huawei Noah's Ark Lab\\
$^2$School of Computer Science, FEIT, University of Sydney, Australia\\
\emails
\{liuchuanjian, yunhe.wang, kai.han, xuchunjing\}@huawei.com,
c.xu@sydney.edu.au
}

\begin{document}

\maketitle

\begin{abstract}
	Exploring deep convolutional neural networks of high efficiency and low memory usage is very essential for a wide variety of machine learning tasks. Most of existing approaches used to accelerate deep models by manipulating parameters or filters without data, e.g., pruning and decomposition. In contrast, we study this problem from a different perspective by respecting the difference between data. An instance-wise feature pruning is developed by identifying informative features for different instances. Specifically, by investigating a feature decay regularization, we expect intermediate feature maps of each instance in deep neural networks to be sparse while preserving the overall network performance. During online inference, subtle features of input images extracted by intermediate layers of a well-trained neural network can be eliminated to accelerate the subsequent calculations.  We further take coefficient of variation as a measure to select the layers that are appropriate for acceleration. Extensive experiments conducted on benchmark datasets and networks demonstrate the effectiveness of the proposed method.
\end{abstract}

\section{Introduction}

\begin{figure}[t]
	\centering
	\setlength{\belowcaptionskip}{-0.2cm}
	\includegraphics[width=\linewidth]{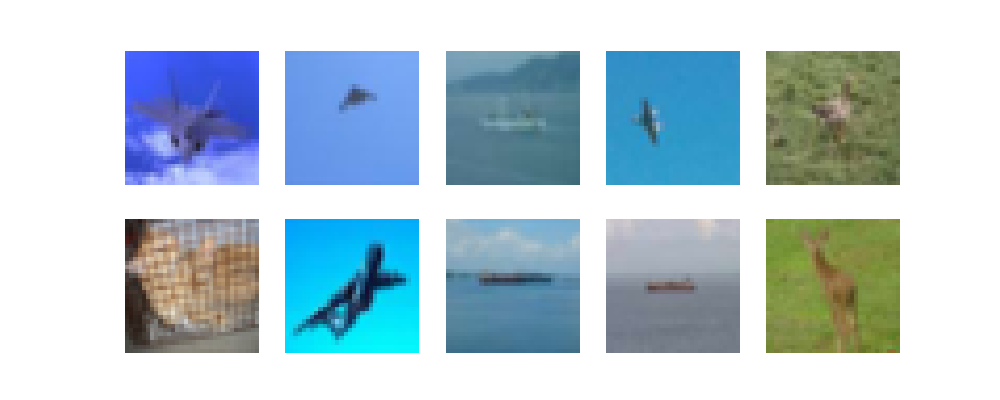}\\
	\small (a) Easy examples with about $65\%$ feature pruning ratio.\\
	\includegraphics[width=\linewidth]{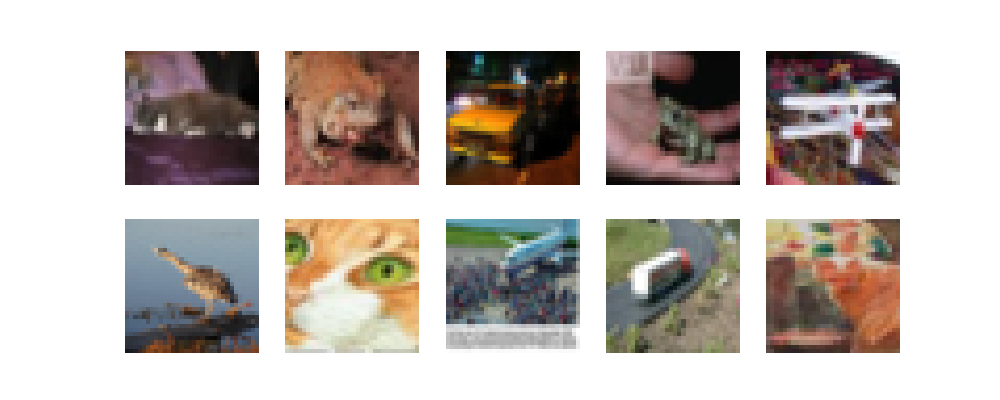}\\
	\small (b) Hard examples with about $40\%$ feature pruning ratio.
	\caption{Examples with different pruning ratios selected using the VGG16 learned on the CIFAR-10 dataset.}
	\label{Fig.1}
\end{figure}

Deep learning methods, especially convolutional neural networks have been successfully applied a number of computer vision tasks, such as image classification~\cite{AlexNet,ResNet}, object detection~\cite{fasterRCNN}, image super-resolution~\cite{VDSR}, \etc. However, most of deep CNNs are well designed with huge parameters and computational complexities for the accuracy reason. For example, VGG16~\cite{VGGnet} needs about $60$ MB memory and $15.4$ GFLOPs (floating-number operations) for a $224\times224$ image. This limits their usage in edge devices, \eg, mobile phones, smart camera. Thus, effective methods for compressing and speeding-up these deep models are urgently required.

To this end, considerable methods have been proposed to reduce memory and FLOPs. ~\cite{quan1} employed vector quantization which makes use of a cluster center to represent similar connections.~\cite{SVD} exploited the singular value decomposition approach and decomposed the weight matrices of fully connect layers. Considering that $32$-bit floating numbers are too fine for CNNs,~\cite{XNORNet} and~\cite{binary2} explored binarized neural networks, whose weights and activations are -1/1.~\cite{pruning} utilized pruning, quantization and Huffman coding together. In addition,~\cite{CNNpackNIPS} introduced the discrete cosine transform (DCT) bases and converted convolution filters into the frequency domain, thereby producing much higher compression ratio and speed improvement. Although these approaches can provide considerable compression and speed-up ratios for the original heavy CNNs, most of them are equally applied on all instances in the train and test set, which ignore the differences between instances.

Actually, natural images are of different complexities for a given neural network, \eg, an image with clean background and moderate objects is easier to be accurately recognized than an image filled with complex textures and multiple objects as shown in Figure~\ref{Fig.1}. To this end, some recent works proposed to explore accelerating methods for different instances, ~\cite{Runtime} embedded a RNN to discard useless channels,~\cite{liu2018dynamic} proposed a multi-branch scheme, \etc. In fact, convolution filters are designed for learning some intrinsic patterns in natural images, \eg, edges, blobs and color, and natural images are combined by these patterns. Thus, discarding same features for all instances is not the optimal solution for reducing the complexity of deep neural networks. Moreover, a higher sparsity ratio is possible for each instance than on entire dataset. Therefore, these methods have a very huge potential for real-world applications.

Although the above mentioned approaches have made tremendous efforts to address the existing problems for applying deep models, most of them discard useless features by introducing additional operations on the output feature maps such as RNN, RL, \etc. Thus, we propose a new framework for learning the instance-wise sparsity in well designed neural networks by exploiting a feature decay regularization. In practice, a feature decay regularization is utilized to make features of different training instances sparse during the training procedure. Then, some features of instances can be discarded without significantly affecting the performance of the resulting network at runtime. Extensive experiments conducted on benchmark models and datasets demonstrate the proposed method can achieve competitive performance but with lower computational complexities.

\section{Related Works}
Conventional model compression and acceleration approaches are designed for removing redundant filters, weights and blocks to obtain compact architectures from pre-trained deep neural networks, which ignore the complexities of different instances in image datasets. 

In order to excavate the complexity of each instance, several works are proposed for assigning different parts of the designed network to different input data dynamically. For example,~\cite{FBS,Runtime,channelgate} utilized attention and gate layers to evaluate each channel and discard some of them with subtle importances during the inference phrase.~\cite{skipnet,spatialadaptive,veit2018convolutional} utilized a gate cell to discard some layers in pre-trained deep neural networks for efficient inference.~\cite{branchynet,liu2018dynamic,decide2017,hydranet,adaptiveneuralnet} further proposed the branch selection operation to allow the learned neural networks to change themselves according to different input data.~\cite{dynamiccapacity,moreisless,ren2018sbnet} applied the dynamic strategy on the activations of feature maps in neural networks.

In summary, most of the above mentioned approaches did not directly process convolution filters or features and introduce extra components, \ie, gates, attention layers, importance predictors and multi-branches to original CNNs, which involves more computations and parameters. In this paper, we attempt to address the dynamic strategy in the feature pruning aspect, and discard redundant features to reduce the computational complexity during the online inference.

\section{Methods}

\begin{figure*}[t]
	\centering
	\setlength{\belowcaptionskip}{-0.2cm}
	\includegraphics[width=\linewidth]{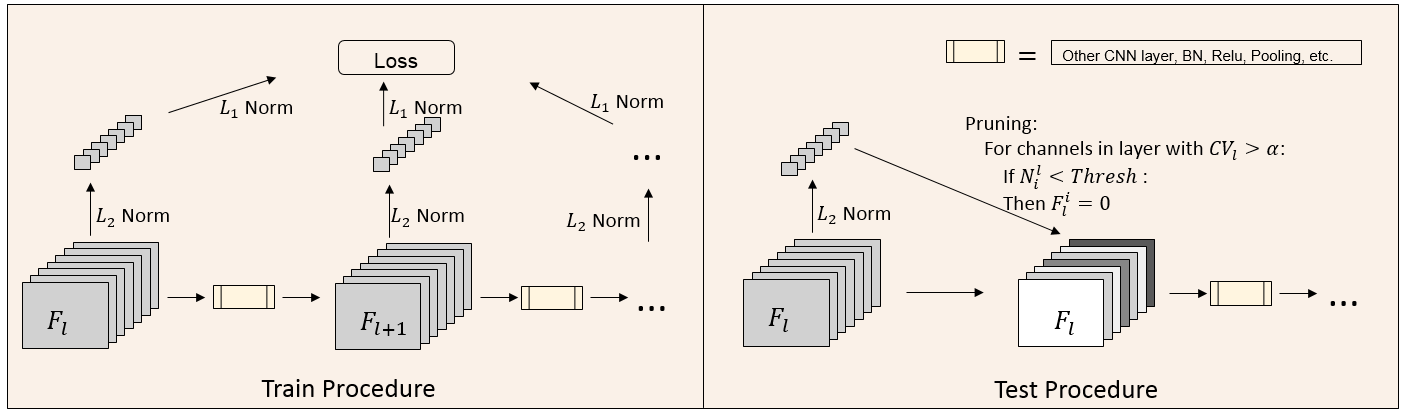}
	\caption{The framework of our methods, includes train procedure with feature regularization and test procedure with feature sparsification.}
	\label{Fig.2}
\end{figure*}

In this section, we investigate the structural sparsity of deep neural networks by a feature decay regularization for obtaining efficient CNN models. In the training procedure, instance-wise $\ell_{2,1}$-norm is included for making features of all instances sparse, while less important features will be eliminated to reduce computational complexity during the inference, as shown in Figure~\ref{Fig.2}.

\subsection{Weight Pruning for Neural Networks}
Given a deep convolutional neural network, denote weights of the $l$-th convolutional layer as a sequence of 4D tensors, \ie, $W^{(l)} \in R^{C_{l+1}\times C_l\times D_l \times K_l}$, where $C_{l+1}$, $C_l$, $D_l$ and $K_l$ are the dimensions of the $l$-th weight tensor along the axes of filter, channel, height and width, respectively. The conventional method for compressing deep neural networks can be formulated as:
\begin{equation}
\min_{W} \mathcal{L}(W, \mathcal{X}, \mathcal{Y}) + \lambda\cdot\sum_{l=1}^L R(W^{(l)}),
\label{eq.1}
\end{equation}
where $\mathcal{L}(\cdot)$ stands for classification loss for training sample $(\mathcal{X}, \mathcal{Y})$, $R(\cdot)$ is the weight regularization item in deep neural networks, and $\lambda$ is a hyper parameter for seeking the trade-off between accuracy and model size.

According to different concerns, the weight regularization $R(\cdot)$ could have various formulations,
\begin{itemize}
	\item[-] $R(W^{(l)})=\sum_{c_{l+1}=1}^{C_{l+1}} \lvert W_{c_{l+1},:,:,:}^{(l)}\rvert_1$ is utilized for removing useless convolution filters~\cite{Thinet};
	\item[-] $R(W^{(l)})=\sum_{c_l=1}^{C_l} \lvert W_{:,c_l,:,:}^{(l)}\rvert_1$ can be applied to eliminate redundant channels~\cite{sparse2};
	\item[-] $R(W^{(l)})=\sum_{d_l=1}^{D_l}\sum_{k_l=1}^{K_l} \lvert W_{:,:,d_l,k_l}^{(l)}\rvert_1$ is often embedded into deep neural networks for discarding subtle weights in CNNs~\cite{pruning}.
\end{itemize}

Wherein, the $\ell_1$-norm applied on each convolution filters, \ie, filter pruning is the most effective scheme for removing useless filters in the pre-trained network and provide a new network with the same number of layers but significantly fewer parameters and FLOPs. However, pruning filters is equal to discard some of output features for all instances, which ignores the difference between complexities of various images. In fact, convolution filters are designed for learning some intrinsic patterns in natural images, \eg, edges, blobs, and color, and images consist of various combinations of these patterns. Thus, discarding all features is not the optimal solution for reducing the complexity of deep neural networks.

\subsection{Instance Feature Sparse Regularization}

As mentioned above, we can obtain a network with fewer convolution filters compared to the original network by minimizing the regularization of convolution filters in pre-trained deep neural networks. In fact, the regularization applied on convolution filters can be shifted to the corresponding features. For an optimal feature elimination, we propose to utilize the $\ell_{2,1}$-norm to learn sparse features of different instances, which is a widely used regularization to select features~\cite{norm210}. In practice, the $\ell_{2,1}$-norm for an arbitrary matrix $M$ with $n$ rows and $m$ columns is defined as:
\begin{equation}
\lvert M\rvert_{2,1} = \sum_{i=1}^{n}\sqrt{\sum_{j=1}^{m}M_{ij}^2},
\label{eq.2}
\end{equation}
which first calculates the $\ell_2$-norm for each row and then stack them by utilizing the $\ell_1$-norm.

Then, we incorporate Eq.\ref{eq.2} in the training procedure of deep neural networks as shown in the left part in Figure~\ref{Fig.2}. Specifically, let $F_{N\times C_l\times H_l\times W_l}^l$ be the features of $l$-th convolutional layer, $N$, $C_l$, $H_l$ and $W_l$ are batch size, channel number, height and width of this layer, respectively. For the $c_l$-th channel of $l$-th layer of $n$-th instance, we compute $\ell_2$-norm for this channel: 
\begin{equation}
N_{n,c_l}^l = \sqrt{\sum_{h_l=1}^{H_l}\sum_{w_l=1}^{W_l}{(F_{n,c_l,h_l,w_l}^l)}^2}.
\label{eq.3}
\end{equation}
For each instance, we then compute its $\ell_1$-norm over all layers in the network: 
\begin{equation}
N_n = \sum_{l=1}^{L}\sum_{c_l=1}^{C_l}N_{n,c_l}^l.
\label{eq.4}
\end{equation}
Thus, the entire objective function of the proposed method can be formulated as
\begin{equation}
\mathcal{L}_{task}+\lambda\cdot\sum_{n=1}^{N}N_n,
\label{eq.5}
\end{equation}
where $\mathcal{L}_{task}$ is final loss function of the given visual task such as classification or detection. $\lambda$ is regularization parameter used for controlling the effect of corresponding feature regularization item. In contrast to weight sparsity, Eq.\ref{eq.5} learns the feature sparsity of each instance in the training procedure simultaneously, which is a more accurate approach for removing redundancy in deep neural networks.

\subsection{Instance-wise Sparsity Inference}
A novel framework for learning instance-wise feature redundancy is proposed in the above section. In order to implement the online inference efficiently, we further explore the testing procedure of the proposed method. The right part of Figure~\ref{Fig.2} illustrates the detailed inference procedure of the proposed method. 
For an arbitrary input image, we first compute the $\ell_2$-norm of feature maps of each convolutional layer using Eq.\ref{eq.3}. 

The feature regularization in training leads to multi-polarization of features, so we utilize the coefficient of variation (CV) to calculate dispersion of norm distribution for each layer in the online inference. It is defined as the ratio of the standard deviation to the mean as follows:
\begin{equation}
\mathcal{CV}_l=\frac{\sqrt{\frac{1}{C_l}\sum_{c_l=1}^{C_l}(N_{n,c_l}^l-\mu_l)^2}}{\mu_l},
\label{eq.7}
\end{equation}
where
\begin{equation}
\mathcal{\mu}_l=\frac{\sum_{c_l=1}^{C_l}N_{n,c_l}^l}{C_l}.
\label{eq.6}
\end{equation}In probability theory and statistics, CV is a standard measure of dispersion of a probability distribution, and it is independent of data scales. The larger CV value of some layer means greater changes in importance of channels, and we tend to drop channels from such layer. We set a global CV threshold as $\alpha$, the layers whose CV values are larger than $\alpha$ could be pruned. 

After determined the prunable layers, we further find the specific channels that could be pruned in these layers. Since the magnitude of feature norms in different layers is various, we cannot use a global threshold to eliminate useless features. Alternatively, we compute the mean $\mu$ as Eq.\ref{eq.6} of $\ell_2$-norms for each layer and set a drop threshold $\beta \in [0,2)$. The channels whose $\ell_2$-norm value is under $\beta\cdot\mu$ are discarded, and only the remained channels will be transmitted into the following layers.

\subsection{Computation Complexity}

Let $F_{N\times C_l\times H_l\times W_l}^l$ be feature map on the $l$-th layer, and the size of filters on the $(l+1)$-th layer is $W_{C_{l+1}\times C_l\times k\times k}$. Standard convolutions have the computation cost of: $C_l\times C_{l+1}\times H_l\times W_l\times k\times k$. The computation cost of $L_2$-norms is $C_l\times H_l\times W_l$, and the computation cost of mean and CV is about $4C_l$. This added computation cost could be ignored in contrast to the convolution. As a result, if we discard $C$ channels for a layer, then we can save $C$/$C_l$ ratio of computation cost approximately at runtime.

\section{Experiments}

In this section, we evaluate feature regularization on general image classification task. Detailed analysis on instance-wise feature sparsity brought by the feature regularization loss are presented. In addition, we demonstrate the instance-wise  channel pruning results and reduced FLOPs.

\subsection{Datasets and Experimental Settings}

We extensively evaluate our methods on two popular classification datasets: CIFAR-10~\cite{cifar10} and Imagenet(ILSVRC2012)~\cite{imagenet}. Three networks are considered: VGG16~\cite{VGGnet}, resnet-18(Res18 for simplicity)~\cite{ResNet} and mobilenet-v2 (MNet2 for simplicity)~\cite{Mobilenet2}. The MNet2 is more challenging to accelerate because of its compactness. Besides, to further verify our method's effectiveness, we did experiments on the network which is compressed by network slimming~\cite{netslim}. 

For VGG16, we substitute the fully connection layer with global average pooling in all experiments. We use filters with size $3\times3$ and stride $1$ on the first convolutional layer for Res18 in CIFAR-10. For MNet2, we reset the strides in first convolutional layer and the second inverted residual block layer to $1$. For preprocessing and other super-parameters in training,  we follow the same routines proposed for these networks.

\subsection{CIFAR-10}

In this section, we make comprehensive experiments and analysis on the instance-wise feature sparsity on CIFAR-10 dataset. A number of $\ell_{2,1}$-norm regularization factors are considered, $\lambda=0$, $1e$-$6$, $1e$-$7$, $1e$-$8$ respectively. When $\lambda=0$, it is equivalent to standard approach. The baseline results are shown in Table~\ref{tab.1}. The feature regularization restricts the feature representation power of deep models, so larger $\lambda$ leads to lower accuracy.

\begin{table}[t]
	\centering
	\setlength{\belowcaptionskip}{-0.2cm}
	\small
	\begin{tabular}{|c|c|c|c|c|}
		\hline
		Network  & $\lambda=0$ & $\lambda=1e$-$8$ & $\lambda=1e$-$7$ & $\lambda=1e$-$6$  \\
		\hline
		VGG16       & 0.942  & 0.939 & 0.934 & 0.560     \\
		Res18       & 0.934  & 0.934 & 0.930 & 0.735      \\
		MNet2       & 0.918  & 0.915 & 0.900 & $-$      \\
		\hline
	\end{tabular}
	\caption{The baseline accuracy of CIFAR-10 test set by different networks and feature regularization factors. '$-$' means no test.}
	\label{tab.1}
\end{table}

\paragraph{Variation of feature distribution.} In our method, CV is a metric measuring the changes in importance of channels. In Table~\ref{tab.2}, we show the changes of the CV values of some convolutional layers. The standard VGG16 has small CV values usually, which means the importance of different channels are similar. In contrast to standard VGG16, we get higher CV values for VGG16 with proposed feature regularization. Additionally, the CV values increase with $\lambda$. With the increase of CV value, the importance of channels produces polarization. In the following, we will explore the trade-off between different CV value thresholds and prune ratio and accuracy.

\begin{figure}[t]
	\centering
	\includegraphics[width=\linewidth]{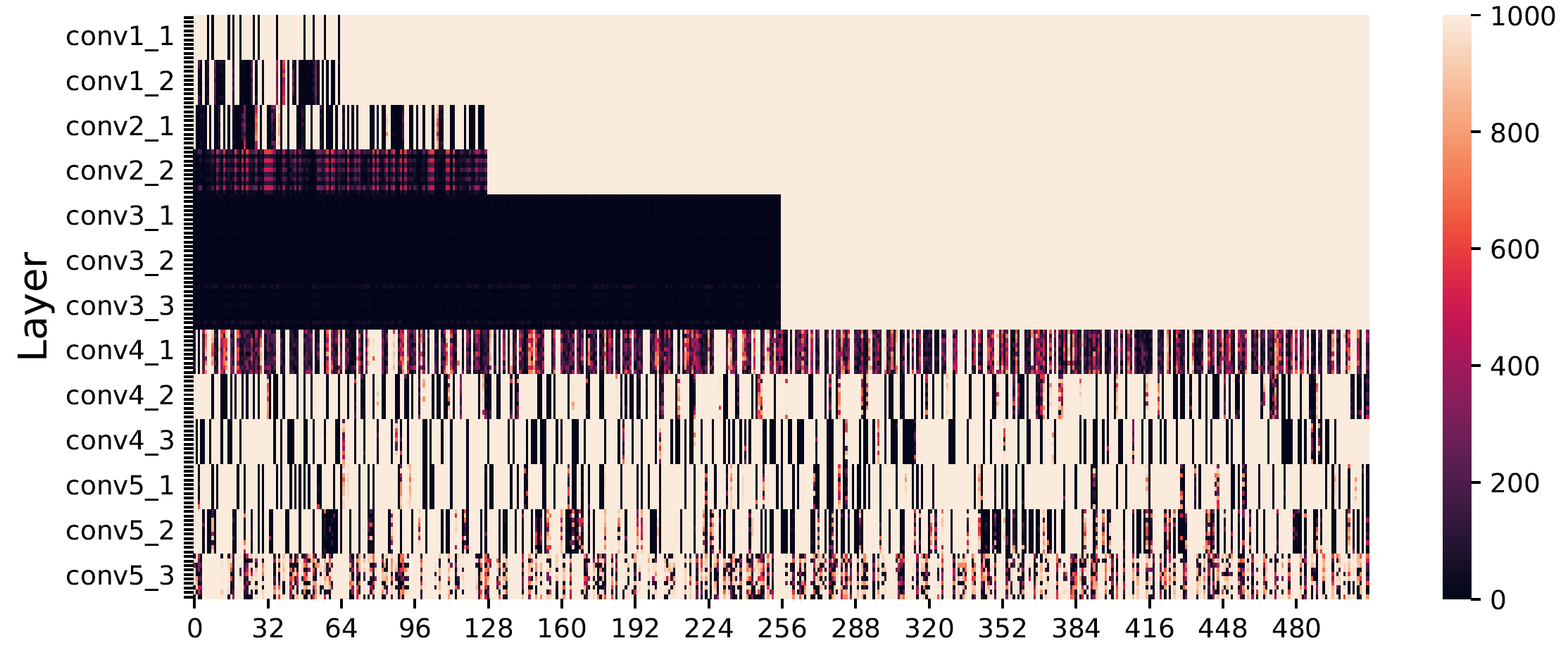}
	\vspace{-1.0em}
	\caption{The channel pruning results of VGG16. The x-axis is the index of channels. The label conv$i_\_j$ in y-axis means the $i_\_j$-th layer, and each layer contains $10$ categories of CIFAR-$10$. Light color means more samples have this channel been dropped, and vice versa. This result is got when $\alpha=0.5$, $\beta=1.0$.}
	\label{Fig.3}
\end{figure}

\begin{table}
	\centering
	\setlength{\belowcaptionskip}{-0.2cm}
	\small
	\begin{tabular}{|c|c|c|c|c|}
		\hline
		$\lambda$   & conv1\_1 & conv3\_1 & conv5\_1 & conv5\_3  \\ 
		\hline
		$0$         & 1.3221   & 0.3412   & 0.7518   & 0.9855    \\
		\hline
		$1e$-$8$    & 1.7955   & 0.3256   & 1.1926   & 1.3256    \\
		\hline
		$1e$-$7$    & 2.2815   & 0.4185   & 2.2749   & 1.9029    \\
		\hline
		$1e$-$6$    & 5.6362   & 6.1605   & 3.3022   & 4.3541    \\
		\hline
	\end{tabular}
	\caption{The mean value of CV for some layers in VGG16.}
	\label{tab.2}
\end{table}

\paragraph{Visualization of pruned channels.} In Figure~\ref{Fig.3} each grid in the heatmap represents the number of samples of one category that discard this channel (there are $1000$ samples for each category in CIFAR-10 test set). The low level layers of VGG16 have the similar channel prune results for different categories. In our opinion, the low layers in CNN usually learn the pixel level information, which are shared by all categories. In the higher layer, more differences of dropped channels presented by this figure. The higher layer has more semantic information of categories, so they tend to discard different channels for different categories. In our model, the middle layers such as conv3$\_{x}$ have merely been pruned due to small CV value. The $10$ easy samples that dropped more channels and $10$ hard samples that dropped less channels during inference are shown in Figure~\ref{Fig.1}. From these examples, we acknowledge that easy samples usually have pure background or texture, moderate size, \etc. The hard samples present complex background and texture, truncated objects, \etc.

\begin{figure}[t]
	\centering
	\setlength{\belowcaptionskip}{-0.4cm}
	\includegraphics[width=\linewidth]{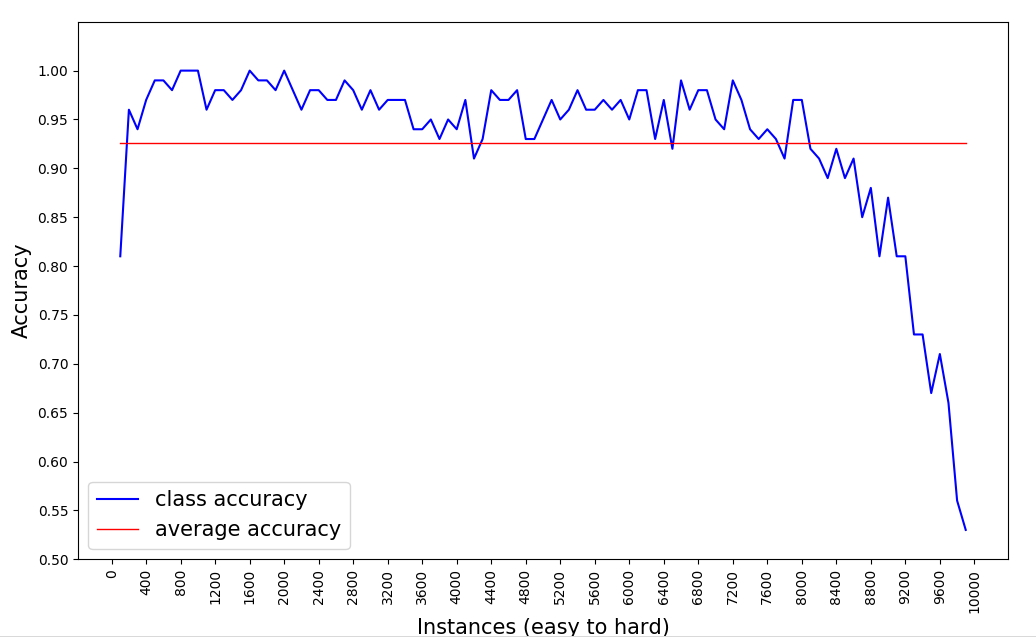}
	\caption{The accuracy of different classes of samples in CIFAR-10 test set. From left to right in x-axis, the number of pruned channel gradually increases. The accuracy is computed every $100$ samples in the blue line. The red line is the global average accuracy across the whole test set.}
	\label{Fig.4}
\end{figure}

\paragraph{Statistics of easy and hard samples.} We group the samples from easy to hard in accordance with the number of pruned channels. From less to more, we compute the accuracy of every $100$ samples by the number of pruned channels for VGG16 in Figure~\ref{Fig.4}. Unusually, the first $100$ easy samples have low accuracy in contrast to the global average accuracy. So we check all these images manually, and find the misclassified samples usually have small object or confusing background. The hard samples have poor accuracy as we believed, and more attention should be paid to these examples to improve accuracy. Besides, we find no significant difference in channel pruning ratio for different categories, the reason lies in that CIFAR-10 has few categories and the test set collects all kinds of instances for each category.

\paragraph{Results of instance-wise feature pruning of VGG16.} We evaluate the proposed instance-wise feature pruning method with random pruning strategy and $\ell_2$-norm based pruning. The random pruning strategy is randomly discarding fixed ratio of channels, while the $\ell_2$-norm based pruning is discarding fixed ratio of channels with minimum $\ell_2$-norm values. Our method is pruning channels that exceed predefined thresholds CV factor $\alpha$ and drop threshold $\beta$. From Table~\ref{tab.3} and \ref{tab.4}, the model with large feature regularization factor $\lambda=1e$-$6$ gets $56.0\%$ accuracy after pruning $95.6\%$ channels for VGG16. The reason lies that large regularization factor limits the representation ability of features, which leads to low accuracy but high feature sparsity. The random method leads to bad accuracy under the same prune ratio, and the norm based method gets relative better results. Our methods has higher accuracy and pruned ratio simultaneously. We use different combinations of CV value $\alpha$ and drop threshold $\beta$. In Table~\ref{tab.4}, for low $\alpha$ and $\beta$, the standard VGG16 gets $94.0\%$ accuracy and $20.7\%$ pruned ratio, and the regularized VGG16 gets $93.6\%$ accuracy and $46.9\%$ prune ratio when $\lambda=1e$-$7$. With low $\alpha$ and high $\beta$, the standard VGG16 gets low accuracy $79.2\%$, and the regularized VGG16 still gets $92.6\%$ accuracy after pruned $53.2\%$ channels. This proves that feature regularization leads to the uneven distribution of features. Then for high $\alpha$ and $\beta$, the standard VGG16 gets $93.8\%$ accuracy but very low prune ratio $4.2\%$, and the regularized VGG16 gets $93.3\%$ accuracy under $45.1\%$ pruned ratio. For all these methods, with the increase of $\lambda$(except $\lambda=1e$-$6$), the test accuracy grows under the same prune ratio. This certificates that large regularization factor $\lambda$ produces large feature sparsity, and more channels could be pruned without sacrificing performance.

\begin{table}
	\centering
	\small
	\setlength{\belowcaptionskip}{-0.1cm}
	\begin{tabular}{|c|c|c|c|c|c|c|}
		\hline
		$pr$   & \multicolumn{2}{|c|}{$10\%$} & \multicolumn{2}{|c|}{$20\%$} & $30\%$ & $40\%$  \\ 
		\hline
		$\lambda$   & rand & min & rand & min & min & min        \\
		\hline
		$0$         & 0.673 & 0.921 & 0.151  & 0.842   & 0.420   & 0.145    \\
		\hline
		$1e$-$8$    & 0.746 & 0.925 & 0.205  & 0.884   & 0.695   & 0.265    \\
		\hline
		$1e$-$7$    & 0.820 & 0.933 & 0.443  & 0.929   & 0.920   & 0.894    \\
		\hline
		$1e$-$6$    & 0.560 & 0.560 & 0.560  & 0.560   & 0.560   & 0.560    \\
		\hline
	\end{tabular}
	\caption{The accuracy after pruning fixed ratio of channels for VGG16 by random and minimum norm value methods on CIFAR-10 test set. The abbreviation 'rand' and 'min' means two pruning method, respectively.}
	\label{tab.3}
\end{table}

\begin{table}
	\centering
	\small
	\setlength{\belowcaptionskip}{-0.3cm}
	\begin{tabular}{|c|c|c|c|c|c|c|}
		\hline
		$thresh$     & \multicolumn{2}{|c|}{$0.5, 0.5$} & \multicolumn{2}{|c|}{$0.5, 1.0$} & \multicolumn{2}{|c|}{$1.0, 1.0$}  \\ 
		\hline
		$\lambda$   & acc & pr & acc & pr  & acc & pr    \\
		\hline
		$0$         & 0.940 & 0.207 & 0.792 & 0.388 & 0.938 & 0.042   \\
		\hline
		$1e$-$8$    & 0.938 & 0.300 & 0.783 & 0.474 & 0.938 & 0.259    \\
		\hline
		$1e$-$7$    & 0.934 & 0.469 & 0.926 & 0.532 & 0.933 & 0.451    \\
		\hline
		$1e$-$6$    & 0.560 & 0.956 & 0.560 & 0.956 & 0.560 & 0.956    \\
		\hline
	\end{tabular}
	\caption{Prune channels with different combinations of threshold $\alpha$ and $\beta$ for VGG16. The abbreviation 'acc' and 'pr' means accuracy and prune ratio, respectively.}
	\label{tab.4}
\end{table}

\paragraph{Results of instance-wise feature pruning of light-weight CNNs.} To further verify the effectiveness of feature regularization, the results on light networks, MNet2 and pruned VGG16 by nets slimming~\cite{netslim}, are shown in Table~\ref{tab.5} and Table~\ref{tab.6}. For MNet2, the average prune ratio is bigger than $0.7$, which suggests that the features in MNet2 are redundant. By check the prune ratio of each layer, we find that the expand layer(expand ratio is $6$) in inverted residual block of MNet2 are pruned much more than the shrink layer, even exceed $95$\% channels are pruned in the higher expand convolutional layers. The depthwise separable convolution itself in standard MNet2 is beneficial to the feature sparsity, which leads to better accuracy and higher prune ratio in contrast to standard VGG16. The regularized MNet2 gains an advantage when CV thresh becomes larger. For pruned VGG16, the baseline of VGG16 with BN scale factor regularization is $0.938$. The accuracy becomes $0.574$ after prune $67\%$ channels with minimum BN scale factor. The we fine-tune this pruned model with proposed feature regularization. In Table~\ref{tab.6}, the second column when $\alpha=0$ and $\beta=0$ is the baseline  accuracy of fine-tuned on pruned VGG16 with different regularization factors. From the last column, we find that another $20.1\%$ or $27.2\%$ computations could be saved when the accuracy drops about $0.8\%$ or $1.7\%$, respectively.

\begin{table}[b]
	\centering
	\small
	\begin{tabular}{|c|c|c|c|c|c|c|}
		\hline
		$thresh$     & \multicolumn{2}{|c|}{$1.5, 1.0$} & \multicolumn{2}{|c|}{$1.25, 0.75$} & \multicolumn{2}{|c|}{$1.0, 0.5$}  \\ 
		\hline
		$\lambda$   & acc & pr & acc & pr  & acc & pr    \\
		\hline
		$0$         & 0.845 & 0.742 & 0.899 & 0.754 & 0.915 & 0.758   \\
		\hline
		$1e$-$8$    & 0.889 & 0.746 & 0.902 & 0.766 & 0.913 & 0.766    \\
		\hline
	\end{tabular}
	\caption{Prune channels with different combinations of thresholds $\alpha$ and $\beta$ for MNet2 on CIFAR-10.}
	\label{tab.5}
\end{table}

\begin{table}
	\centering
	\small
	\begin{tabular}{|c|c|c|c|c|c|c|}
		\hline
		$thresh$     & \multicolumn{2}{|c|}{$0, 0$} & \multicolumn{2}{|c|}{$0.5, 1.0$} & \multicolumn{2}{|c|}{$0.7, 0.7$}  \\ 
		\hline
		$\lambda$   & acc & pr & acc & pr  & acc & pr    \\
		\hline
		$0$         & 0.938 & 0 & 0.414 & 0.287 & 0.638 & 0.232    \\
		\hline
		$1e$-$7$    & 0.937 & 0 & 0.891 & 0.203 & 0.930 & 0.201    \\
		\hline
		$1e$-$6$    & 0.925 & 0 & 0.921 & 0.203 & 0.921 & 0.272    \\
		\hline
	\end{tabular}
	\caption{Prune channels with different combinations of thresholds $\alpha$ and $\beta$ for pruned VGG16 by net slimming on CIFAR-10.}
	\label{tab.6}
\end{table}

\begin{table}[t]
	\centering
	\small
	\setlength{\belowcaptionskip}{-0.1cm}
	\begin{threeparttable}
		\begin{tabular}{|c|c|c|c|c|}
			\hline
			Methods   & Network  & Error  & prune ratio & FLOPs($10^8$)  \\ 
			\hline
			LCCL      &  Res20   & 8.32   & 65.1\tnote{1} & 0.26    \\
			\hline
			Gating    & Res18    & 6.00   & 73.1 & 1.35   \\
			\hline 
			FBS       & M-Cifar\tnote{2}  & 8.45 & 30.0 & 1.7   \\
			\hline
			\multirow{2}{*}{Ours}    & VGG16    & 6.61   & 46.9 & 1.84     \\ \cline{2-5}  
			& Res18    & 6.63   & 40.3 & 0.98    \\ \cline{2-5}
			& MNet2    & 8.72   & 76.6 & 0.31     \\
			\hline
		\end{tabular}
		\begin{tablenotes}
			\footnotesize
			\item[1]{Ratio of zero elements, not channels.}
			\item[2]{This is computed according to the $2\times$ speed-up ratio in paper.}
		\end{tablenotes}
	\end{threeparttable}
	\caption{Comparisons of error rate and prune ratio between different methods. Gating uses large batch size(256) and more epochs(300). FBS constructs 8-layer M-CifarNet with 1.3M parameters.}
	\label{tab.7}
\end{table}

\paragraph{Comparative analysis.}  We compare our methods with several dynamic accelerate methods. ~\cite{moreisless} used low-cost collaborative layer(LCCL) to pre-compute the non-zero elements in the out feature maps, then ignored the locations with zeros. Channel gating(Gating)~\cite{channelgate} worked by identifying the unimportant channels and turning off them. \cite{FBS} proposed feature boosting and suppression, a new method to predictively amplify salient convolutional channels and skip unimportant ones at run-time. As shown in Table~\ref{tab.7}, the feature regularization method provides a better trade-off between error and FLOPs reduction under the similar network architecture. The feature regularization method is easy to implement in contrast to other methods which need extra network structures. These dynamic accelerated methods are complementary to static methods such as pruning, quantification and low rank decomposition. It is possible to obtain better accuracy and FLOPs reduction by combining these methods to statical models.

\subsection{Imagenet}

ImageNet is a large dataset, which has $1.3$M images and $1000$ categories. Due to the diversity of categories and richness of images, more features are expected to be utilized by deep neural networks in order to get higher discriminability. 

\begin{figure}[htb]
	\centering
	\includegraphics[width=\linewidth]{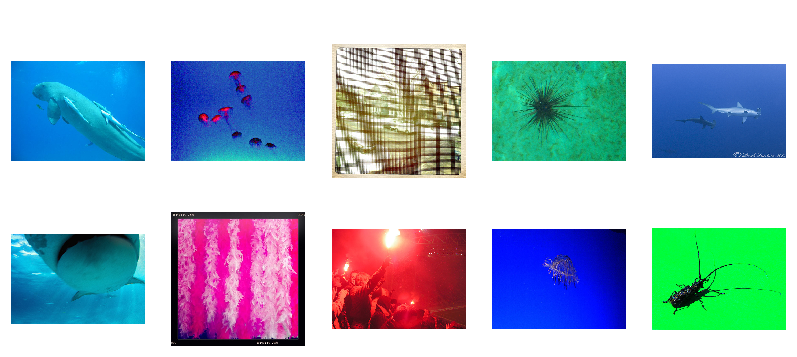}\\
	\small (a) Easy examples selected by VGG16 on Imagenet.\\
	\includegraphics[width=\linewidth]{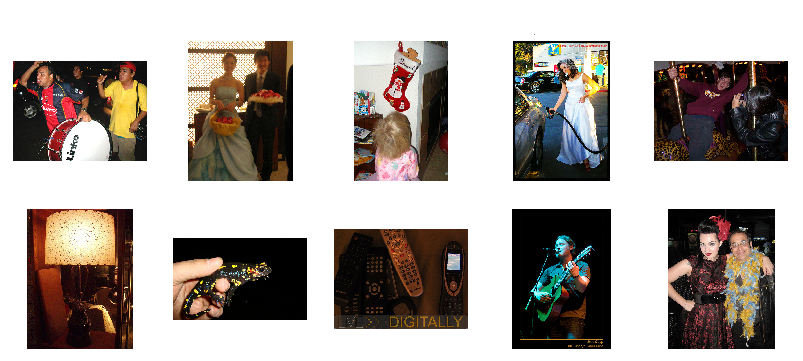}\\
	\small (b) Hard examples selected by VGG16 on Imagenet.\\
	\caption{The easy and hard samples selected from Imagenet with inference feature pruning. The easy samples have more channels been pruned, and hard samples have less channels been discarded.}
	\label{Fig.5}
\end{figure}

Figure~\ref{Fig.5} presents the easy and hard samples of Imagenet. There are more categories in Imagenet in contrast to CIFAR-10, the skew of channel prune ratio for different categories emerged. Some categories like 'window screen', 'space bar' and 'window shade' have more channels been dropped, these categories can be regarded as easy categories. The hard categories contains 'butcher shop', 'jinrikisha', 'oxcart', etc. 

\begin{table}
	\centering
	\small
	\begin{tabular}{|c|c|c|c|c|c|c|}
		\hline
		$thresh$     & \multicolumn{2}{c|}{$0, 0$} & \multicolumn{2}{c|}{$1.0, 0.3$} & \multicolumn{2}{c|}{$0.8, 0.3$}\\ 
		\hline
		$\lambda$   & acc & pr & acc & pr  & acc & pr   \\
		\hline
		$0$         & 0.706 & 0 & 0.702 & 0.335  & 0.694 & 0.385   \\
		\hline
		$1e$-$8$    & 0.701 & 0 & 0.698 & 0.439  & 0.692 & 0.475   \\
		\hline
	\end{tabular}
	\caption{Prune channels with different combinations of thresholds $\alpha$ and $\beta$ for MNet2 on Imagenet.}
	\label{tab.8}
\end{table}

From Table~\ref{tab.8}, some interesting results are found. Firstly, the MNet2 has near half of channels been dropped in spite of its compactness, which is less than the pruned channels in CIFAR-10. The inverted residual block in MNet2 expands the channels for the inputs initially, then shrink channels in the last convolutional layer of this block. Detailed analysis on the dropped channels show that almost all the dropped channels locate in the expand layer, which means that one small expand ratio in inverted residual block may get equivalent accuracy with large expand ratio. The result is consistent with CIFAR-10. Secondly, we find the standard MNet2 also has better accuracy under relatively lower prune ratio in contrast feature regularization method. But in the similar accuracy $69.4\%$ and $69.2\%$, feature regularized model has more channels $47.5\%$ to $38.5\%$ been pruned.

\section{Conclusion}

The differences between instances make the dynamical pruning become possible. This paper introduces structural feature regularization on CNN models in the training procedure to get instance-wise sparsity, and prune useless channels during inference for each instance. The proposed method is not designed for reducing disk memory occupation and GPU space. In fact, we tend to accelerate the inference by selectively computing only a subset of channels that are important for the input. As a side effect, the amount of cached activations and the number of read, write and arithmetic operations would be largely decreased, which leads to the improvement of memory occupation and computation at runtime. Within $1\%$ accuracy loss, we demonstrate that VGG16 can drop $50\%$ channels on CIFAR-10, and MNet2 can drop $47\%$ channels on Imagenet. In addition, the inference channel pruning method can be easily combined with other existing static network compression methods to get better speed-up ratio. Further to mine the restrictions on features maps to improve efficiency or performance of networks will be studied.

\section*{Acknowledgments}

Chang Xu was supported by the Australian Research Council under Project DE180101438.

{\clearpage

	\bibliographystyle{named}
	\bibliography{ijcai19}
}

\end{document}